\documentclass[a4paper,english,submission]{rnti}

\usepackage[T1]{fontenc}
\usepackage[utf8]{inputenc}

\usepackage{url}

\usepackage{graphicx}
\usepackage{adjustbox}

\usepackage{booktabs, multirow}
\usepackage{makecell}
\usepackage{float}
\usepackage{subfig}
\usepackage{bbold}
\newcolumntype{?}{!{\vrule width 2pt}}

\titrecourt{Comparing facial emotion recognition methods}

\nomcourt{R. El Cheikh et al.}

\titre{A comparative study of emotion recognition methods using facial expressions}

\auteur{Rim El Cheikh\affil{1},
        Hélène Tran\affil{1}\\
        Issam Falih\affil{1},
         Engelbert Mephu Nguifo\affil{1}}

\affiliation{
    \affil{1}Université Clermont-Auvergne, CNRS, Mines de Saint-Étienne, \\Clermont-Auvergne-INP, LIMOS, 63000 Clermont-Ferrand, France\\
            \{rim.el\_cheikh,helene.tran\}@doctorant.uca.fr \\
            \{issam.falih,engelbert.mephu\_nguifo\}@uca.fr    
 }

\resume{%
Understanding the facial expressions of our interlocutor is important to enrich the communication and to give it a depth that goes beyond the explicitly expressed. In fact, studying one's facial expression gives insight into their hidden emotion state. However, even as humans, and despite our empathy and familiarity with the human emotional experience, we are only able to guess what the other might be feeling.
In the fields of artificial intelligence and computer vision, Facial Emotion Recognition (FER) is a topic that is still in full growth mostly with the advancement of deep learning approaches and the improvement of data collection.
The main purpose of this paper is to compare the performance of three state-of-the-art networks, each having their own approach to improve on FER tasks, on three FER datasets. The first and second sections respectively describe the three datasets and the three studied network architectures designed for an FER task. The experimental protocol, the results and their interpretation are outlined in the remaining sections.%
}

\summary{%
Comprendre les expressions faciales de notre interlocuteur est important pour enrichir la communication et lui donner une profondeur qui va au-delà de l'explicitement exprimé. En fait, étudier l'expression du visage donne un aperçu de l'état émotionnel caché. Cependant, même en tant qu'êtres humains, et malgré notre empathie et familiarité avec l'expérience émotionnelle humaine, nous ne pouvons que deviner ce qu'une personne pourrait ressentir. Dans les domaines de l'intelligence artificielle et de la vision par ordinateur, la reconnaissance des émotions faciales (FER) est un sujet qui est encore en pleine croissance, principalement avec l'avancement des approches d'apprentissage en profondeur et l'amélioration de collecte de données. L'objectif principal de cet article est de comparer les performances de trois réseaux de l'état de l'art, chacun ayant sa propre approche pour améliorer les tâches FER, sur trois jeux de données FER. Les première et deuxième sections décrivent respectivement les trois jeux de données et les trois architectures de réseaux étudiées pour effectuer une tâche FER. Le protocole expérimental, les résultats et leur interprétation sont décrites dans les sections restantes.%
}

\begin{document}

%
\section{Introduction}

Automatic Emotion Recognition is an area of research that has been active for many decades. Researchers are not only trying to mimic human logical-mathematical intelligence but have been for many years now trying to explore emotional functions.
However, despite many impressive results in terms of predictive ability, automatic methods still have a long way to go when it comes to offering a satisfying model that can capture the complexity, heterogeneity, and subjectivity of the emotional experience.

The interest in this topic stems from different concerns. FER has applications in various fields. For instance, given that understanding the emotional state of our interlocutor is essential for a deep and rich communication, recognizing emotions from facial expressions could be seen as a wish to further develop the human-machine interaction. Also, having an automatic FER that offers reliable results on complex emotional states could be a breakthrough in the fields of healthcare \citep{Kashif} and psychology \citep{Tadalagi}.  

A key element for building a robust machine learning model is the availability and quality of data. In the context of emotion recognition using images of facial expressions, collecting a big amount of data is a manageable task as proven by the multitude of available FER datasets. The samples in these datasets are collected from different sources. For example, images of faces could be collected from the internet and can be considered as expressions captured "in-the-wild". The samples could be pre-processed to only keep the facial region~\citep{Barsoum,Mollahosseini} or we can choose to also incorporate and analyze the context in which the image is captured ~\citep{Kosti}. Other datasets contain images taken in the lab, where subjects are asked to act the facial expressions in a controlled environment ~\citep{Jaffe, Lucey}. Other differences between datasets could come from the existence of noisy data, the lighting, the encoding of the samples, the diversity of the faces in terms of age, gender, race, ethnicity, etc. It is also important to insist of the subjectivity in expressing and identifying emotions, which introduces bias and ambiguity \citep{Tran} in the annotation step, as well as during the interpretation of the results step. In fact, one way to try to build a more accurate and precise emotion recognition model is to combine different types of data that could be informative on the emotional state of a person, such as facial expression, speech, tone, body posture, physiological signal, etc. In this situation, processing and combining data from different sources creates an additional challenge to the design of a robust model. 

As a first step to building such systems, we choose to focus this work on methods that predict emotions from facial expression images only. Three state-of-the-art models for FER tasks have been selected for experiments. These models diverge not only in their architectures but also in the level on which they intervene to better the prediction quality. Our experiments provide a fair comparison of their performance on three datasets that differ in terms of their size, the setting in which the samples are captured, and the distribution of classes. 
The paper starts by giving an overview of three existing facial expression datasets used for emotion prediction and that were chosen for our experiments: FER+, AffectNet and CK+. Then, three neural network architectures are described in section 2, followed by a presentation of the experimental protocol in section 3. In sections 4 and 5, results are respectively presented and analyzed. The code used for the experiments is publicly available at \url{https://anonymous.4open.science/r/FE-rec-0F0B/}.

\section{Dataset Description}
\label{db_sec}
This section describes the three datasets, FER+, AffectNet and CK+, used for our experiments.

\subsection{FER+}
FER+, introduced by ~\cite{Barsoum}, is one of the most known and used datasets for FER tasks. FER+ is annotated with 8 emotions: Neutral, Happiness, Surprise, Sadness, Anger, Disgust, Fear, and Contempt, as well as Unknown and NF (Not a Face). 
The images are collected from the internet and contain only the facial region. They are gray-scale and are of size 48x48 pixels, which could be considered as very low resolution. To avoid affecting the quality of predictions, we resized the images to 96x96 pixels.

FER+ can be considered as an improvement of the FER-2013 dataset, which was introduced by ~\cite{Goodfellow}. While the labeling for FER-2013 was done by two persons only (the authors of the dataset), FER+ took advantage of the increasingly popular scheme of crowdsourcing in order to collect ground truth labels and the images of FER-2013 were relabeled  by 10 crowdsourced taggers. Each image has an emotion probability distribution by using the annotations given by the 10 taggers. The FER+ dataset is therefore a multi-labeled set of images that better reflects the ambiguity and diversity of facial emotions. In our single-label classification experiments, the label affected to a sample is the one with the highest number of votes. If the label with the highest number of votes is Unknown or NF, we choose to exclude the sample from the experiments.

\subsection{AffectNet}
The AffectNet dataset~\citep{Mollahosseini} provides 291,651 images of faces annotated with 8 categories of emotions: Neutral, Happy, Sad, Surprise, Fear, Disgust, Anger and Contempt. A larger version of AffectNet provides annotations such as None, Uncertain, and No-Face that were not taken into account in the implementation but are important to ensure the quality of the annotated dataset. 
The images present in AffectNet were collected from the internet by the means of queries such as "joyful girl", "astonished senior", etc. Only the face region was kept and 12 annotators were asked to annotate the samples.
The authors provide the agreement percentages between two annotators over the annotations. It is interesting to note that all classes resulted in agreements ranging from 50\% to 70\%, except Happy with 79.6\%. These percentages are similar to the accuracy reportedly found by recognition models in the FER literature that are trained on this dataset~\citep{Siqueira,Farzaneh}.

The images are in RGB and of size 224x224 pixels. In addition to the discrete emotion categories, AffectNet gives a continuous annotation of the faces in a two dimensional space, valence-arousal. The valence dimension is an indicator for the pleasantness of the emotion and arousal is a measure of the emotion intensity. These annotations were not used for our experiments, but provide precious information that enables us to extend our research to emotion recognition using continuous modeling.

\subsection{CK+}
The Extended Cohn-Kanade, or CK+, dataset~\citep{Lucey}, contains 593 sequences of facial expressions captured from 123 subjects in a lab. A sequence starts with a neutral expression and ends with the peak expression where the facial Action Units (AUs) are coded. Action Units were proposed by ~\cite{Ekman} as a way to model facial expressions by encoding the movements of facial muscles. 
From these 593 sequences, only 327 are labeled with a category of emotions: Anger, Contempt, Disgust, Fear, Happy, Sadness and Surprise. The unlabeled sequences are considered as non fitting for the prototypical definition of the emotions taken into account and are not used for supervised training. The labeling is done by assigning an emotion to the facial expression if one or more AUs are detected, with respect to the Facial Action Coding System manual ~\citep{Ekman}. 

Given the small number of available sequences, we chose to take three images from each sequence instead of only the peak expression. This way, the number of samples is increased for each class of emotions. Moreover, the neutral class is created by taking the first frame of each sequence, so that the set of emotions used for CK+ is the same as the other two datasets. The images are sized either 640x490 or 640x480 pixels. Some were gray-scale while others were RGB. In our experiments, the images were fed to the networks as gray-scale 640x490 pixel arrays.

\section{Studied networks}
This section describes the methods chosen for experiments in order to compare their performance on the previously presented datasets. The choice was based on the fact that each method provides a different approach in terms of improving on facial emotion recognition tasks. Availability of implementations of these models was also an important factor in choosing them.

\subsection{ESR}
In the paper titled “Efficient Facial Feature Learning with Wide Ensemble-based Convolutional Neural Networks”, by ~\cite{Siqueira} introduced the Ensembles with Shared Representations (ESR) network.
Ensemble learning combines the predictions obtained by different classifiers that are usually trained in an individual and independent manner, hence producing a more accurate output than a single model would. 
The ESR network makes use of the translation-invariance property of the patterns learned by each convolutional layer.
In fact, the patterns that are learned in the early layers of the network (low-level features) can be considered in some way as common to all the images that the network might encounter. These patterns can be oriented lines, edges, or colors. 
As we go deeper into the layers, the patterns that the network is learning become more complex and specific to each image. These features, in the case of facial expressions, could be the shape of the eyes, of the mouth, and of the nose.
Therefore, in ESR we find two main building pieces:
\begin{enumerate}
    \item The base of the network:
            A line-up of convolutional layers that are responsible for learning the low-level features. As mentioned before, patterns learned at this level of the network are very general and therefore can be shared by multiple branches. This base uses a transfer learning mechanism that speeds up the learning process as the ensemble grows, while improving the performance, since the best configuration from the training of each branch is reloaded as base.
    \item The independent convolutional branches:
            A branching of the convolutional layers allows each branch to learn its own individual high-level features.
\end{enumerate}

Finally, the optimization of the network consists in minimizing a loss function that combines the loss function of each branch.

In the case of in-the-lab databases, the ESR network contains only 4 independent convolutional branches, as input data is expected to be of good quality (good lighting, adjusted head pose, etc.). By contrast, when dealing with in-the-wild datasets, the number of branches is increased to 9.

\subsection{SCN}
In the Self-Cure Network~\citep{Wang}, the authors addressed the problem of uncertainty that comes with the labeling of the facial expression datasets.
One of the reasons behind is the problem of subjectiveness when categorizing the human emotional experience.
Furthermore, in datasets where images are captured in-the-wild, the uncontrolled setting is definitely a source of inconsistency and uncertainty.
To overcome this problem, the SCN network proposes a relabeling step before the recognition task to perform robust feature learning with uncertainty.
First, the features are extracted from the images using a “backbone” convolutional network, which can be any traditional CNNs. In the implementation provided by the authors and as well as in our experiments, the backbone used is ResNet-18~\citep{He} pre-trained on the MS-Celeb-1M face recognition dataset~\citep{Guo}.
Second, if a sample is considered uncertain, it is relabeled.
The relabeling can be summarized in three steps:
\begin{enumerate}
    \item Self-attention importance weighting: 
    Importance weights are computed for each sample using a linear fully-connected layer with a sigmoid activation function. The assigned importance weight reflects the contribution that each image has on the classifier training.

    \item Ranking regularization:
     The weights are regularized in order to reduce the importance given to uncertain samples. The samples are ranked by importance and two groups of samples are created: the low-importance (30\%) and the high-importance images (70\%).  
    
    \item Relabeling: This module assigns a new emotion label (the one with highest predicted probability produced by the model) to the images from the low-importance group. For that purpose, softmax probabilities are used to determine which image is actually incorrectly labeled. If the difference between the predicted probability for the label initially given to the sample, and the highest predicted probability produced by the model, is higher than a given threshold (set to 0.2 by the authors), then the data is relabeled according to the highest probability found by the model. 
    
\end{enumerate}

\subsection{DACL}
DACL, or Deep Attentive Center Loss~\citep{Farzaneh}, is a facial expression recognition method that uses an attention mechanism to estimate attention weights that are correlated with feature importance. In fact, in the training phase, learning irrelevant features is harmful for the performance of the network. Therefore, the authors proposed the integration of a Deep Metric Learning (DML) approach that enhances the learning of discriminative features by the model.

The first step is to feed a convolutional neural network with the input images in order to generate the feature maps, followed by the DACL component composed of two building pieces:
\begin{enumerate}

    \item Context Encoder Unit:
        This generates latent representations for each spatial feature map that is outputted by the backbone CNN. All these feature maps represent the context, and therefore the obtained latent feature vector is reduced in dimension and is devoid of the noise, only containing relevant information from the initial features extracted by the CNN. 
        The linear layer weights were initialized according to the Kaiming initialization~\citep{He_delving}, which is appropriate when using the ReLU activation function, as it helps to capture non-linear relationships between layers.
    
    \item Multi-head binary classification module:
    This component estimates the attention weights from the latent representation outputted by the Context Encoder Unit.
    At this point, the problem is considered as a multi-label classification problem, where an attention weight (softmax) is computed at each component for the latent feature vector.
\end{enumerate}

Finally, these attention weights are used to compute the sparse center loss, which in turn is fractionally used, alongside the loss computed by the CNN backbone to compute the final loss. The authors empirically gave 1\% of the final loss to the sparse center loss.

\section{Experimental Setting}
In order to fairly compare the performance of the three networks mentioned above on the FER+, AffectNet and CK+ datasets, we followed an identical experimental protocol for training and testing them.
A 5-fold cross-validation is used to evaluate the model. 
In fact, the three datasets are constructed in different ways: FER+ has training, validation and testing subsets, AffectNet has training and validation subsets, and CK+ has no subsets. Therefore, in the case of FER+ and AffectNet the subsets are mixed before performing cross-validation.
The dataset is split as follows: 80\% for training, 10\% for validation, and 10\% for test. Table \ref{tab:dbs} shows the distribution of the eight emotions for the three datasets. 
The training is performed over 60 epochs, with Adam optimizer and a 0.001 learning rate.

\begin{table}[ht]
\centering
\begin{adjustbox}{width=\columnwidth,center}
    \begin{tabular}{|c | c c c  c | c c c c | c c c  c|}\hline 
        &\multicolumn{4}{c|}{\textbf{FER+}}  &\multicolumn{4}{c|}{\textbf{AffectNet}} &\multicolumn{4}{c|}{\textbf{CK+}} \\\cline{2-13}
        &Train &Valid &Test &Total &Train &Valid &Test &Total &Train &Valid &Test &Total \\\hline
        Neutral &8796 &1148 &1052 &10996 &3737 &464 &471 &4672 &262 &28 &37 &327 \\
        Happy &7231 &912 &896 &9039 &6283 &756 &815 &7854 &166 &16 &25 &207 \\
        Sad &3001 &351 &399 &3751 &1509 &192 &185 &1886 &67 &9 &8 &84 \\
        Surprise &3153 &378 &410 &3941 &1043 &147 &113 &1303 &199 &31 &19 &249 \\
        Fear &546 &67 &69 &682 &691 &81 &92 &864 &60 &6 &9 &75 \\
        Disgust &126 &15 &17 &158 &587 &72 &75 &734 &141 &23 &13 &177 \\
        Anger &2125 &253 &278 &2656 &1449 &188 &174 &1811 &108 &14 &13 &135 \\
        Contempt &120 &13 &17 &150 &563 &83 &58 &704 &43 &4 &7 &54 \\ \hline
        Total &25098 &3137 &3138 &31373 &15862 &1983 &1983 &19828 &1046 &131 &131 &1308 \\
        \hline
    \end{tabular}
\end{adjustbox}
\caption{Class distribution in the three studied dataset.}\label{tab:dbs}
\end{table}

\subsection {Evaluation Metrics}
\label{metrics}
This subsection presents the metrics used for the model evaluation, we define TP as the value of true positives and FP is the false positives. 

\paragraph{Accuracy and Balanced Accuracy}
The overall accuracy is reported in our results. However, it does not take into account the highly imbalanced label distribution found in the three datasets.
For this reason, we report the balanced accuracy defined as the average of recall (see further in this section) obtained on each class. This definition proposed by~\cite{Mosley} is equivalent to the most commonly used formula for accuracy where each sample is weighted by the prevalence of its true label.
The chosen formula computes an aggregated score of the measurements of the predictive quality for each class independently. 

\paragraph{Precision and Recall}
For our experiments, we compute precision score, which measures the predictor's ability not to label a negative sample as positive, weighted by the support of the labels. Similarly, the weighted recall score is computed by averaging the recall score for each label after weighting it by the support of the label.
\paragraph{F1-score}
We also report the F1-score as the harmonic mean of the weighted precision and the weighted recall.
\paragraph{AUC ROC}
The AUC ROC score reflects the discrimination ability of the classifier between the different classes.
In our experiments, we report the prevalence-weighted average of AUC ROC scores computed for each class against all the others.

\section {Results}
In this section we present the results of the five-fold cross-validation. First, we evaluate and compare the models on the whole dataset in terms of predictive performance. Second, the models are compared by taking into account their capability to discriminate each emotion class. 

\subsection{Overall classification}
Table \ref{tab:res} reports the computed metrics on the test subsets averaged over the 5 folds. This fairly quantifies the predictive performance of the models while taking into account the imbalance in the datasets.
When trained on FER+, DACL shows a better accuracy than ESR and SCN. Weighted accuracy is decreased compared to overall accuracy, which is expected given the imbalance in the classes distribution in FER+. 
Also, DACL shows a better precision and recall compared to the other models, as well as a higher ability to discriminate between labels (AUC ROC).
For AffectNet, DACL also shows the best performance in terms of recall, precision and AUC ROC scores. However, the best balanced accuracy is achieved by ESR while staying relatively close to the balanced accuracy of DACL.
Regarding CK+, ESR shows the best performance with a really high accuracy of  91.5\% and a very satisfying balanced accuracy compared to the other models. The F1-score is also the highest for ESR on this dataset. 
However, SCN produces the best AUC ROC score which shows that it is the best model in terms of discrimination between the categories. 

\begin{table}[ht]
 \begin{center}
   \begin{tabular}{|c | c | c | c | c | c | c | c |} 
         \hline 
            &  & acc & bal acc & pr & rec & F1 & AUC ROC \\ [0.5ex] 
            \hline   
            \multirow {6}{*}{FER+} & {ESR} & \makecell{0.857 \\ ± 0.013}
                                            & \makecell{0.617 \\ ± 0.029}
                                            & \makecell{0.855 \\ ± 0.015}
                                            & \makecell{0.857 \\ ± 0.013}
                                            & \makecell{0.856 \\ ± 0.014}
                                            & \makecell{0.937 \\ ± 0.023} \\\cline{2-8}
                                & {SCN} & \makecell{0.810 \\ ± 0.012}
                                            & \makecell{0.520 \\ ± 0.028}
                                            & \makecell{0.808 \\ ± 0.010}
                                            & \makecell{0.810 \\ ± 0.012}
                                            & \makecell{0.809 \\ ± 0.011}
                                            & \makecell{0.956 \\ ± 0.002}\\\cline{2-8}
                                & {DACL} & \makecell{\textbf{0.867} \\ ± 0.005}
                                            & \makecell{\textbf{0.647} \\ ± 0.018}
                                            & \makecell{\textbf{0.863} \\ ± 0.005}
                                            & \makecell{\textbf{0.867} \\ ± 0.005}
                                            & \makecell{\textbf{0.865} \\ ± 0.005}
                                            & \makecell{\textbf{0.973} \\ ± 0.002}\\\cline{2-8}
                                            
            \hline
            \multirow {6}{*}{AffectNet} & {ESR} & \makecell{0.648 \\ ± 0.002} 
                                                & \makecell{\textbf{0.439} \\ ± 0.001}
                                                & \makecell{0.626 \\ ± 0.006} 
                                                & \makecell{0.648 \\ ± 0.002} 
                                                & \makecell{0.637 \\ ± 0.004} 
                                                & \makecell{0.821 \\ ± 0.002} \\\cline{2-8}
                                        & {SCN} & \makecell{0.651 \\ ± 0.002}
                                                    & \makecell{0.390 \\ ± 0.011}
                                                    & \makecell{0.622 \\ ± 0.005}
                                                    & \makecell{0.651 \\ ± 0.002}
                                                    & \makecell{0.636 \\ ± 0.003}
                                                    & \makecell{0.894 \\ ± 0.004}\\\cline{2-8}
                                        & {DACL} & \makecell{\textbf{0.664} \\ ± 0.016}
                                                    & \makecell{0.429 \\± 0.033} 
                                                    & \makecell{\textbf{0.633} \\ ± 0.017}
                                                    & \makecell{\textbf{0.664} \\ ± 0.016}
                                                    & \makecell{\textbf{0.648} \\ ± 0.017}
                                                    & \makecell{\textbf{0.901} \\ ± 0.007}\\\cline{2-8}
            \hline
            \multirow {6}{*}{CK+} & {ESR} & \makecell{\textbf{0.915} \\ ± 0.018}
                                            & \makecell{\textbf{0.888} \\ ± 0.034}
                                            & \makecell{\textbf{0.922} \\ ± 0.014}
                                            & \makecell{\textbf{0.915} \\ ± 0.018}
                                            & \makecell{\textbf{0.918} \\ ± 0.016}
                                            & \makecell{0.945 \\ ± 0.008} \\\cline{2-8}
                                & {SCN} & \makecell{0.820 \\ ± 0.030}
                                            & \makecell{0.703 \\ ± 0.072}
                                            & \makecell{0.798 \\ ± 0.043}
                                            & \makecell{0.820 \\ ± 0.030}
                                            & \makecell{0.808 \\ ± 0.032}
                                            & \makecell{\textbf{0.962} \\ ± 0.010} \\\cline{2-8}
                                & {DACL} & \makecell{0.846 \\ ± 0.125}
                                            & \makecell{0.790 \\ ± 0.175}
                                            & \makecell{0.843 \\ ± 0.154}
                                            & \makecell{0.846 \\ ± 0.125}
                                            & \makecell{0.844 \\ ± 0.140}
                                            & \makecell{0.951 \\ ± 0.051}\\\cline{2-8}                                        
         \hline
\end{tabular}
\caption{Average performance metrics computed on the test subsets afetr the five-fold cross-validation. The value after (±) symbol represents the standard deviation across five folds. "acc" is accuracy, "bal acc" is balanced accuracy, "pr" is weighted precision, "rec" is weighted recall, "F1" is the F1-score, "AUC ROC" is the Area under the ROC Curve.}\label{tab:res}
 \end{center}
\end{table}

\subsection{Emotion-specific classification}
\begin{figure}[H]%
\makebox[\textwidth][c]{
    \subfloat
        [\footnotesize \centering Confusion matrices of ESR predictions on FER+(left), AffectNet(middle) and CK+(right)]
            {{\includegraphics[width=.4\textwidth]{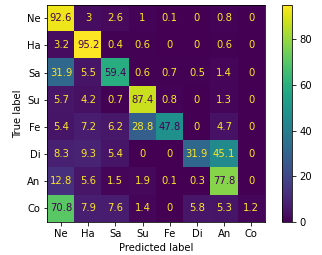}}
             {\includegraphics[width=.4\textwidth]{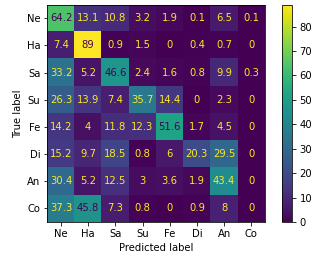}}
             {\includegraphics[width=.4\textwidth]{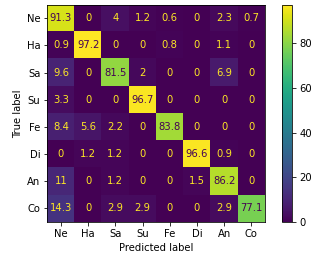}}
            }%
        }
    \makebox[\textwidth][c]{
    \subfloat
        [\footnotesize \centering Confusion matrices of SCN predictions on FER+(left), AffectNet(middle) and CK+(right)]
            {{\includegraphics[width=.4\textwidth]{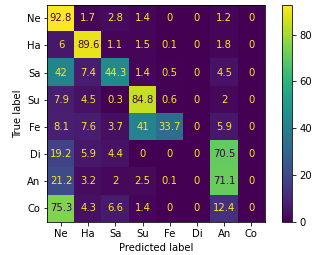}}
            {\includegraphics[width=.4\textwidth]{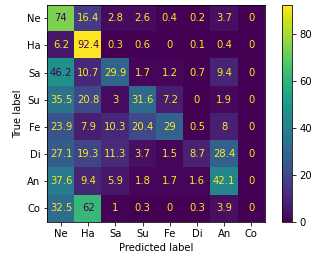}}
             {\includegraphics[width=.4\textwidth]{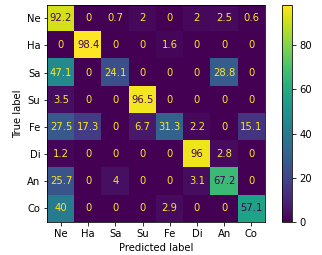}}
            }%
    }
    \makebox[\textwidth][c]{
    \subfloat
        [\footnotesize \centering Confusion matrices of DACL predictions on FER+(left), AffectNet(middle) and CK+(right)]
            {{\includegraphics[width=.4\textwidth]{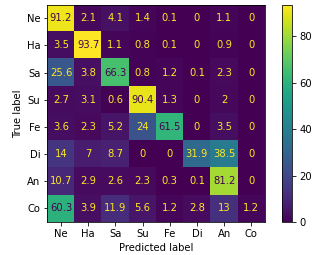}}
            {\includegraphics[width=.4\textwidth]{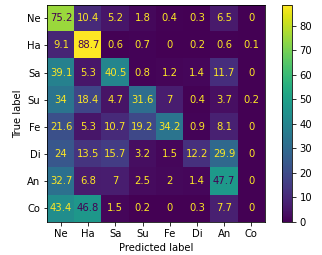}}
             {\includegraphics[width=.4\textwidth]{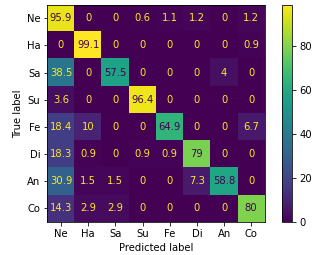}}
            }%
    }
    \caption{Normalized confusion matrices, each computed by averaging the confusion matrices obtained over the 5 folds. }
    \label{fig:conf_mats}%
\end{figure}

In order to have a closer look on the performance of each model for each of the 8 emotion categories, normalized confusion matrices are shown in figure \ref{fig:conf_mats}. 
When training on FER+, we find a higher TP value for the classes that are over-represented. For example, a high percentage of images that are labeled "Happy" in the dataset were correctly classified. However, not more than 1.2\% of "Contempt" images were predicted as such by the three models. We can clearly see that for FER+ the imbalance in the training sets impacts the learning process by the models. 
Also, "Contempt" expressions are more often than not misclassified as "Neutral" by all three of the models. 
When training on AffectNet, we can see that "Neutral" and "Happy" have a high classification accuracy for the three models. This is not the case for all other classes, which have a high probability of being misclassified as "Neutral" (and as "Happy" for "Contempt" samples). 
As the distribution classes in AffectNet shows, "Happy" has a higher frequency than "Neutral" in AffectNet, this was translated by a higher positive values for "Happy" than for "Neutral".  
Finally, when training on CK+, ESR predicts correct labels for over 80\% of the samples for all classes except for "Contempt" which is accurately predicted in 77.1\% of cases. However, this is not the case for SCN and DACL, where a high accuracy is only found for classes that are over-represented in CK+ (Happy, Neutral and Surprise). This said, all three models trained on CK+ still manage to provide a better accuracy on almost all under-represented classes compared to the other two datasets.

\section {Discussion}
\texttt{DACL} provides the best scores when training on FER+ and \texttt{AffectNet}. Both these datasets contain facial expressions captured in-the-wild, where the subjects expressed their emotions in a natural way. This is why samples from FER+ and \texttt{AffectNet} might require a representation that takes into account the characteristics of the whole face region. It could be said that the Context Encoder unit of DACL is helping to represent the overall context from the face region of these samples and the multi-head attention mechanism allow the network to retain the most important information that are useful to infer the emotion.
ESR gives the best accuracy and F1-score for CK+. Therefore, we can say that the exploitation of shared representations that is characteristic of ESR, proves effective when the samples are taken in a controlled environment and when the expressions are posed and intentional.  

In figure \ref{fig:conf_mats}, it is noteworthy that training on FER+ induces a lot of mistakes by predicting "neutral" for samples that are labeled "sad" (using ESR, 59.4\% of "sad" are correctly predicted and 31.9\% are predicted as "neutral"). This is not the case for samples that are labeled "surprise" (using ESR, 87.4\% of "surprise" are correctly classified), despite the fact that both classes, "sad" and "surprise" are represented in the dataset in very close proportions (3751 and 3941 samples respectively). This raises the question of what is it about "sad" expressions that are less discernible than "surprise" expressions.  
However, both ESR and DACL manage to correctly classify 31.9\% of the "disgust" samples, which is an under-represented class. This is not the case for SCN. In fact, we can see that SCN's characteristic of dealing with uncertainty in the annotations of the training set does not translate well on the predictions done on a test set. 
As for training on AffectNet, the classification of a higher number of samples from other classes into "neutral" than into "happy" is proof that this misclassification comes from the ambiguity of emotions and not necessarily from the distribution in the dataset. We can say that "neutral" is a safe choice to categorize an emotion when we are somewhat uncertain. 
Similar to AffectNet, we find that when SCN and DACL are trained on CK+, a high number of misclassified samples are in fact predicted as "neutral". This shows that even a controlled manner of capturing emotions induces the mistake of safely labelling an emotion as "neutral". 

In our experiments, we chose to compare the performance of three different neural networks, \texttt{ESR}, \texttt{SCN} and \texttt{DACL}. They all have different architectures and different approaches to deal with FER tasks. However, they cannot represent the whole class of methods to which they belong (attention-based, dealing with uncertainty, or ensemble learning). Other architectures for each method class~\citep{Li,Hao} could be studied in order to have a more comprehensive view of their performance in the context of FER applications. The same could be said about datasets, where we can extend our experiments to other datasets~\citep{Jaffe,Li_raf} that share characteristics with the ones we discussed in this paper.

\section {Conclusion}
This paper compares the performance of three neural networks that have different approaches to tackle FER challenges: the first uses ensemble learning and transfer learning to learn facial emotion patterns, the second addresses the problem of subjectivity and uncertainty when labelling emotions, and the third makes use of an attention mechanism to learn from relevant features. To evaluate these models, we used three datasets that are different in terms of data collection and capture setting.
FER+ and AffectNet are both in-the-wild datasets with facial expressions captured in an uncontrolled setting.
The model that uses an attention mechanism provides the best results on images that are captured in the wild, which was expected as this type of images is very noisy and it would have been difficult to recognize the emotion if the model was not guided in focusing on the relevant parts of the images. On CK+, containing images with posed expressions, the model based on ensemble and transfer learning is the one that performs the best: the network is build to exploit low-level features, which is suitable in a setting where noisy parameters are mitigated (face in front of the camera with a neutral background).
Moreover, our experiments show that models often mistake some emotion classes (e.g. "contempt" and "neutral") in in-the-wild datasets, showing that emotion ambiguity alters the model's discrimination abilities. 
In our comparative study, many challenges were identified, such as the under-represented emotion classes found in many FER datasets and the ambiguity of facial expressions. Also, extending the experiments to more models and datasets would provide a reliable benchmark to choose an adapted FER model in terms of the application at hand.

\bibliographystyle{rnti}
\bibliography{biblio}

\end{document}